\documentclass{article}
\usepackage{iclr2027_conference,times}
\usepackage{preamble}

\setlist{compact}

\providecommand{\vivek}[2][]{{\protect\color{violet}{[\textbf{Vivek\if\relax#1\relax\else(#1)\fi:} #2]}}}

\title{Measuring Empowerment from Offline Data for Intrinsically Motivated Exploration}
\date{March 3, 2026}

\author{%
Ishir Garg\textsuperscript{1}, Vivek Myers\textsuperscript{1}, Qiyang Li\textsuperscript{1}, \\
\textbf{Catherine Ji\textsuperscript{2}, Benjamin Eysenbach\textsuperscript{2} \& Sergey Levine\textsuperscript{1}} \\
\textsuperscript{1}UC Berkeley \quad \textsuperscript{2}Princeton University \\
{\small\texttt{\{ishirgarg,vmyers\}@berkeley.edu}, \texttt{\{qcli,svlevine\}@eecs.berkeley.edu}} \\
{\small\texttt{\{cj7280,eysenbach\}@princeton.edu}}
}

\iclrfinalcopy %

\begin{document}
\maketitle
\lhead{Preprint}

\begin{abstract}
    Large, unlabeled datasets of prior experience have become increasingly available for agentic tasks in domains such as robotics and coding.
A common recipe for these tasks is to train a policy using the prior data, and then further improve it with online rollouts.
But while much work has studied how to use both the offline and online data for policy extraction, far less work has been done to understand how the prior data can be used to efficiently \textit{explore} during the online learning phase.
Previous work has attempted to formalize this idea by defining the \textit{empowerment}, a measure of how many future states an agent can realize; states with higher empowerment require more exploration as there are more futures to explore.
However, existing methods for estimating empowerment require online policy rollouts in the environment.
We propose the first method for estimating empowerment from unlabeled prior data, without any environment interaction.
By jointly learning an empowerment-maximizing policy together with its discounted state-occupancy measure, our estimator can be trained on
unlabeled trajectories generated from any unknown policy. We demonstrate that our empowerment estimator qualitatively gives intuitive empowerment measures and show that the pretrained estimator can be used to improve exploration in the online goal-conditioned reinforcement learning setting.
Code is available at \url{https://github.com/ishirgarg/offline-empowerment}.

\end{abstract}

\section{Introduction}
\label{sec:introduction}

Online exploration is a challenging yet central problem in reinforcement learning, where the agent must traverse the environment to seek out novel states with potentially high reward values.
Previous methods tend to place equal weights across all states in the prior, where the agent is optimized to visit all of these states until all of them have been visited.
But without any prior, the agent must visit all states, increasing the sample-complexity.
This is suboptimal as some states naturally grant the agent more control over what it could do (e.g., picking up a key lets the agent open a door to a new room).

A natural way to characterize the idea above is to consider empowerment~\citep{emps}, a metric that measures to what degree an agent can influence its future states via its actions. States with high empowerment have many distinct future outcomes that the agent can achieve by taking distinct sequences of actions. Because these high-empowerment states have more future outcomes, it seems reasonable to explore more around these states as an agent has the potential to discover more future outcomes or behaviors. Conversely, in low-empowerment states, there is very little that an agent can do to influence its future outcome, so it should spend less time exploring these regions of the state space as there is less strategic value to be gained.

Previous works have explored using empowerment, or related information-theoretic objectives for online exploration~\citep{off-dads, kim-emi, kumar-mi-explore, becker-emp-gain, infopower}, but operate in the purely online setting, where we do not have access to prior data.
Luckily, in many practical environments, we have prior data in the form of unlabeled trajectories.
Ideally, we would like to use the prior data to inform us how to explore online, but estimating empowerment purely from offline data is difficult: existing empowerment estimators~\citep{mohamed-rezende,vic,diayn,dads,off-dads,kumar-mi-explore,becker-emp-gain,infopower,kim-emi} rely on online samples where a skill-conditioned agent must interact with the environment during the optimization.
Furthermore, adapting online methods to the offline setting by simply regularizing the policy as is typical in the offline RL literature~\citep{kumar2020conservative, td3bc, iql} does not work (Section~\ref{sec:baseline-emp-maps}).
Existing empowerment estimators look at whether the outcomes achieved by a policy align with its intentions; however, offline, we do not know the policy's intention.
\textbf{How can we learn empowerment from offline data to improve online exploration?}

In this work, we provide a novel algorithm for learning empowerment purely from \textit{offline data}, without additional environment interaction. Our key contribution is an offline estimator of empowerment: by learning an empowerment-maximizing policy in tandem with its discounted state occupancy, we can extract an estimator of the empowerment from prior data alone. We show that our method produces qualitatively accurate empowerment maps, correctly identifying which states admit the most diverse future outcomes. We further demonstrate that the offline-extracted empowerment estimate can be used to improve exploration in online tasks. Specifically, we find that using the empowerment as an auxiliary loss term improves exploration on downstream, online tasks in OGBench environments \citep{ogbench}. We evaluate our methods in sparse-reward, goal-conditioned environments, where knowing \textit{where} to explore is harder due to the sparsity of the reward, and the agent must prioritize which states to visit. In this setting, our empowerment estimator provides a principled way to characterize the exploration value of a state from offline data, directing the agent toward regions with greater potential for diverse future behaviors.

\section{Related Work}

\paragraph{Empowerment.}

Empowerment~\citep{emps} measures an agent's control over its future as the mutual information between its action sequence (or skill choice) and the resulting future state.
Prior works have estimated empowerment from \textbf{online} data by optimizing a variational lower bound~\citep{mohamed-rezende}, and later work estimates the empowerment of an agent's \emph{current} policy from off-policy samples, by maximizing lower bounds on the empowerment~\citep{kumar-mi-explore, misc, belghazi-mine, myers2024learninga, infopower, choi2021variationalempowermentrepresentationlearning, zheng2025misl}.
However, these methods estimate the empowerment of the current goal-reaching policy rather than the maximum achievable empowerment, which can be a very weak bound in practice~\citep{poole2019variational}.
\citet{becker-emp-gain} compute empowerment directly, but require finite action spaces and incur cost exponential in the horizon.
\citet{pretrain_emp} show that empowerment pretraining provides a strong initialization for downstream finetuning, but use a fixed estimator constructed by a human annotator.
The closest to our work is EELMA \citep{eelma} which estimates the \textit{effective} empowerment (defined in \cref{sec:prelims}) using contrastive learning for discrete LLM actions from collected trajectories; we adapt this method to the continuous action space setting and use it as a baseline in \cref{sec:baseline-emp-maps}.
Critically, none of these empowerment estimators work well when adapted to the offline RL setting (\cref{sec:baseline-emp-maps}).
\textbf{Our work is the first to estimate the maximal empowerment purely from unlabeled offline data.}

\paragraph{Intrinsic motivation and curiosity-driven exploration.} A large body of work in reinforcement learning has studied intrinsic motivation as a mechanism for encouraging exploration in sparse-reward environments.
Early approaches include count-based exploration and density modeling methods that assign bonuses to rarely visited states~\citep{tang2017exploration}.
More recent methods such as curiosity-driven learning estimate prediction error or learning progress in a forward dynamics model as an intrinsic reward signal~\citep{rnd, kim-emi}.
Our work builds on these ideas, using empowerment as an exploration bonus that can efficiently leverage prior data.

\paragraph{Exploration from offline data.} Offline RL learns policies purely from fixed datasets without further environment interaction.
Offline-to-online methods pretrain a policy or value function on the offline data and then fine-tune it with online interaction~\citep{jsrl, calql}.
Prior methods have shown that an online RL agent can be trained to explore much more efficiently in the environment by leveraging offline data~\citep{explore, supe}, but do not define a principled way to characterize the ``exploration value'' of a state.
Our work instead learns empowerment offline as a principled exploration objective.

\section{Preliminaries}
\label{sec:prelims}
We formulate the environment as a Markov Decision Process (MDP) defined by a state space $\mathcal{S}$, action space $\mathcal{A}$, transition dynamics $P(s'|s,a)$, and a discount factor $\gamma$. Let trajectories be denoted by the random variables $s_0, s_1, \dots, s_T$. We are particularly interested in the distribution of discounted future states, denoted as $s^+ \triangleq s_{t+k}$ where $k \in \{0, 1, 2, \dots\}$ is drawn with probability $(1-\gamma)\gamma^{k}$~\citep{crl}. Let $\pi(a \mid s, z) \in \Pi^z$ be a skill-conditioned policy, where $z \in \mathcal{Z} = \{1, \dots, K\}$ is a latent skill drawn from a fixed, pre-determined set of $K$ skills and represented as a one-hot vector. In line with previous work~\citep{successor-representation}, we define the discounted state occupancy measures as
\begin{align}
p^z(s^+ \mid s,a)
&\triangleq
\mathbb{E}_{\substack{
s_{t+1} \sim P(\cdot \mid s_t,a_t), \\
a_t \sim \pi(\cdot \mid s_t, z)
}}
\left[
\sum_{k=0}^{\infty} (1-\gamma)\,\gamma^{k}
\;\mathbf{1}\big(s_k = s^+\big)
 \middle| s_0 = s, a_0 = a\right],
\\
p^z(s^+ \mid s)
&\triangleq
\mathbb{E}_{a \sim \pi(\cdot \mid s, z)}
\left[
p^z(s^+ \mid s, a)
\right].
\end{align}

\textbf{Learning occupancy measures by temporal differences.} Because $s^+$ is
drawn from a geometric mixture over future time steps, the occupancy measures
satisfy a Bellman-style recursion~\citep{successor-representation}:
\begin{align}
p^z(s^+ \mid s, a) &= (1-\gamma)\,\mathbf{1}\big(s^+ = s\big) \;+\; \gamma\,\mathbb{E}_{s' \sim P(\cdot \mid s, a)}\!\big[p^z(s^+ \mid s')\big], \label{eq:occ_bellman_q}\\
p^z(s^+ \mid s) &= \mathbb{E}_{a \sim \pi(\cdot \mid s, z)}\!\big[p^z(s^+ \mid s, a)\big]. \label{eq:occ_bellman_v}
\end{align}
This recursion can be used as a temporal-difference target to learn occupancy
measures from off-policy transitions $(s,a,s')$, without rollouts of $\pi$
itself. By analogy with value functions, we write $Q^z(s^+ \mid s,a)$ and $V^z(s^+ \mid s)$ for learned models of $p^z(s^+ \mid s,a)$ and $p^z(s^+ \mid s)$, respectively.

\textit{Empowerment}~\citep{emps, mohamed-rezende} quantifies the maximal influence an agent can exert over its future outcomes from a given state, typically measured as the channel capacity between a control or intention and a future outcome. Prior work has offered many variations of the specific form. We use the definition from~\citet{vic}, which defines the empowerment in state $s$ as the maximum possible mutual information between a uniformly-distributed random latent skill $z \in \mathcal{Z} = \{1, \dots, K\}$ and the resulting future state $s^+$, conditioned on the current state:
\begin{equation}
\mathcal{E}(s) \triangleq \max_{\pi\in\Pi^{z}} I^{\pi}(s^{+} ; z \mid s_t = s). \label{eq:potential_emp}
\end{equation}
We call this the \emph{potential} or \emph{maximal empowerment}, to distinguish it from the \emph{effective empowerment} defined next. Intuitively, empowerment should be high when a policy can realize a diverse set of outcomes where each outcome corresponds to a distinct skill. For example, an agent stuck in the dead end of a maze has very low empowerment, as its actions have almost no outcome.

We take $\Pi^z$ to be the class of Gaussian policies with a learned mean $\mu_\theta(s,z)$ and a fixed standard deviation $\sigma_a$, clipped to the action bounds: $a = \mathrm{clip}(\mu_\theta(s,z) + \eta,\, -1,\, 1)$ with $\eta \sim \mathcal{N}(0, \sigma_a^2 I)$. For a deterministic policy in a deterministic environment, each occupancy measure $p^z(\cdot \mid s)$ is a set of point masses on the states of a single trajectory, so it has no density over the continuous state space, and the mutual information theoretically saturates: $I^\pi(s^+; z \mid s) = \gamma \log K$ whenever no two skills' trajectories share a state after $s$, no matter how close together they are. With fixed action noise, each skill's occupancy measure is spread over a neighborhood of its trajectory instead, so the skills' occupancy measures overlap and the mutual information reflects how far apart the skills' futures are.

While potential empowerment represents the theoretical maximum control, evaluating it directly would require searching over all policies. Instead, we can define the \textit{effective empowerment} as a lower bound induced by a specific, known policy $\pi$:
\begin{equation}
\tilde{\mathcal{E}}(s;\pi) \triangleq I^\pi(s^+ ; z \mid s_t = s)
\;=\; \frac{1}{K} \sum_{z \in \mathcal{Z}} \mathbb{E}_{s^+ \sim p^z(\cdot \mid s)}\!\left[ \log \frac{p^z(s^+ \mid s)}{\frac{1}{K} \sum_{z' \in \mathcal{Z}} p^{z'}(s^+ \mid s)} \right],
\label{eq:effective_emp}
\end{equation}
where the equality uses that $z$ is uniform; $\tilde{\mathcal{E}}(s;\pi) \le \mathcal{E}(s)$, with equality for the maximizing $\pi$.

\textbf{Our method iteratively discovers a policy to maximize this effective empowerment}, thereby approximating the potential empowerment $\mathcal{E}(s)$, using samples $(s, a, s')$ from an offline dataset $\mathcal{D}$.

\textbf{Problem formulation.}
\label{sec:problem}
We assume access only to a static dataset $\mathcal{D}$ of transitions $(s, a, s')$ collected by an unknown behavior policy $\pi_\beta$, with no further environment interaction. Our goal is to use $\mathcal{D}$ to obtain an estimator $\hat{\mathcal{E}}(s)$ of the potential empowerment $\mathcal{E}(s)$ (Eq.~\ref{eq:potential_emp}) at any state $s$.

\section{An Algorithm for Offline Empowerment}
\label{sec:background}

In this section, we present our algorithm for estimating empowerment from
offline data. The key insight is that the effective empowerment of a policy
depends on that policy only through its discounted state occupancy measures
(Eq.~\ref{eq:effective_emp}), which can be learned off-policy using
temporal differences (Eqs.~\ref{eq:occ_bellman_q}--\ref{eq:occ_bellman_v}). Our
method jointly learns \textbf{occupancy models} $Q^z_\theta(s^+ \mid s,a)$ and $V^z_\theta(s^+ \mid s)$ for a skill-conditioned policy by TD on offline transitions (Section~\ref{sec:occupancy}), trains that \textbf{policy} $\pi(a \mid s, z)$ to maximize the effective empowerment computed from these models so that it approaches the potential empowerment $\mathcal{E}(s)$ (Section~\ref{sec:policy-opt}), and reads off an \textbf{empowerment estimator} $\hat{\mathcal{E}}(s)$ from these models (Section~\ref{sec:estimator}). Throughout training, we sample skills $z \sim p(z)$ uniformly from $\mathcal{Z}$.

\subsection{Learning the Occupancy Measure}
\label{sec:occupancy}

To evaluate effective empowerment offline (\cref{sec:problem}), we must
estimate the distribution of future states without environment interactions. We
achieve this by learning function approximators $Q_\theta$ and $V_\theta$ that
model the discounted future state occupancy measures so that
$Q^z_\theta(s^+ | s, a) \approx p^z(s^+ | s, a)$ and
$V^z_\theta(s^+ | s) \approx p^z(s^+ | s)$. Each training example consists of a
transition $(s, a, s') \sim \mathcal{D}$ together with a random state
$s^+$ drawn uniformly from all states in $\mathcal{D}$, independently of the
transition. Because the empowerment-maximizing policy differs from the behavior
policy, these sampled futures are not technically samples from $p^z(\cdot \mid s, a)$, but the recursive backup allows us to still learn these quantities.

\textbf{Bellman Backups.}
We train the occupancy networks by using the recursion of \cref{eq:occ_bellman_q}--\ref{eq:occ_bellman_v} as a TD target, with the next state $s'$ sampled from the dataset and the expectation over actions in \cref{eq:occ_bellman_v} taken under the current policy $\pi(\cdot \mid s, z)$.
A caveat is that the first term of \cref{eq:occ_bellman_q}, $(1-\gamma)\mathbf{1}(s^+ = s)$, is almost never observed, as a random state $s^+$ drawn from the dataset almost surely differs from the current state $s$. We resolve this by adding a second loss where $s^+ = s$ instead of the randomly sampled offline data state, so that its target includes the $(1-\gamma)$ term.
Equivalently, we sum two regression losses with equal weight,
\begin{align}
    \mathcal{L}^{\mathrm{boot}}_Q & \;=\; \mathbb{E}_{(s,a,s', s^+)}\!\Big[\,\ell\big(\,Q^{z}_\theta(s^+ \mid s, a),\;\, \gamma\,V^{z}_{\bar\theta}(s^+ \mid s')\,\big)\Big],           \\
    \mathcal{L}^{\mathrm{self}}_Q & \;=\; \mathbb{E}_{(s,a,s',s^+)}\!\Big[\,\ell\big(\,Q^{z}_\theta(s \mid s, a),\;\, (1-\gamma) + \gamma\,V^{z}_{\bar\theta}(s \mid s')\,\big)\Big],
\end{align}
with total objective $\mathcal{L}_Q = \mathcal{L}^{\mathrm{boot}}_Q + \mathcal{L}^{\mathrm{self}}_Q$, where $\ell$ is a clipped linear exponential regression loss used in prior works for stability~\citep{linex} (Section~\ref{sec:additional-details}). The V-loss is
\begin{align}
    \mathcal{L}_V & \;=\; \mathbb{E}_{(s,a,s', s^+)}\!\Big[\,\ell\big(\,V^{z}_\theta(s^+ \mid s),\;\, \mathbb{E}_{a' \sim \pi(\cdot \mid s, z)}\big[Q^{z}_{\bar\theta}(s^+ \mid s, a')\big]\,\big)\Big]
\end{align}
Note that $Q_{\bar\theta}, V_{\bar\theta}$ denote target networks for Q and V. We update the networks using Polyak averaging~\citep{ddpg} with an update rate of $\tau=0.005$.

\textbf{Latent Space Parameterization.}
Recall that computing empowerment (\cref{eq:effective_emp}) requires samples from $p^z(s^+ \mid s)$. We now describe a parameterization of $V_\theta$ that makes such sampling tractable.

To tractably model these high-dimensional distributions, we project states and actions into a shared representation space of dimension $d$. We define non-linear encoders $\phi_Q(s, a, z)$, $\phi_V(s, z)$, and $\psi(s^+)$, and parameterize the log-occupancies as scaled negative Euclidean distances:
\begin{equation}
    \log Q^z(s^+ | s,a) = -\tfrac{1}{d} \|\phi_Q(s,a,z) - \psi(s^+)\|^2, \quad
    \log V^z(s^+ | s) = -\tfrac{1}{d} \|\phi_V(s,z) - \psi(s^+)\|^2
    \label{eq:v_latent_param}
\end{equation}
This formulation allows us to sample directly from the discounted state-occupancy measure as an isotropic Gaussian in the latent space $p(\psi \mid \phi) = \mathcal{N}(\phi, \frac{d}{2}I)$, which is needed to compute the empowerment. Note that while we cannot sample $s^+$, we can still sample $\psi(s^+)$ which is sufficient for computing $V^z(s^+ \mid s)$ as we can plug the sampled latent into \cref{eq:v_latent_param}.

It is possible to use separate networks for both $\phi_Q, \phi_V$, or alternatively, only learn $Q^z(s^+ \mid s, a)$ and simply use  $V^z(s^+ \mid s) = \mathbb{E}_{a \sim \pi(\cdot \mid s, z)}\big[Q^z(s^+ \mid s, a)\big]$. We find that using separate networks for Q and V leads to more stable training and qualitatively better estimates of the empowerment.

\subsection{Policy Optimization via the Empowerment Gradient}
\label{sec:policy-opt}
Now that we can learn and sample from the state occupancies, we optimize a policy $\pi(a \mid s, z)$ to maximize its effective empowerment. Because empowerment depends only on the state, we need a way to describe how a skill's action distribution influences it. We can define $\mathcal{M}^{z'}(s^+ \mid s,\pi,z)$ as the marginal modified discounted state occupancy measure, representing the distribution over $s^+$ if skill $z$ draws its action at $s$ from $\pi(\cdot \mid s, z)$ and all skills follow the current policy thereafter:
\begin{align}
    \mathcal{M}^{z'}(s^+ \mid s,\pi,z) = \mathbf{1}\{z' = z\}\,\mathbb{E}_{a \sim \pi(\cdot \mid s, z)}\big[Q^{z'}(s^+ \mid s, a)\big] + \mathbf{1}\{z' \neq z\}V^{z'}(s^+ \mid s).
\end{align}

We can then define the $(\pi,z)$-modified empowerment,
\begin{equation}
    \mathcal{E}_\Delta(s; \pi, z) \;=\; \frac{1}{K} \sum_{z' \in \mathcal{Z}} \mathbb{E}_{s^+ \sim \mathcal{M}^{z'}(\cdot \mid s, \pi, z)} \left[ \log\!\left( \frac{\mathcal{M}^{z'}(s^+ \mid s, \pi, z)}{\frac{1}{K} \sum_{z'' \in \mathcal{Z}} \mathcal{M}^{z''}(s^+ \mid s, \pi, z)} \right) \right],
\end{equation}
which quantifies the local change in empowerment caused by changing the action distribution of skill $z$ at $s$. Increasing this encourages the skill-conditioned policy to take actions that make its future state occupancies under different skills distinguishable. Since we can easily sample from the marginal occupancy measure $V^{z'}$ via our Gaussian parameterization from above, we rewrite this as an importance-weighted expectation and estimate it by Monte Carlo sampling:
\begin{equation}
    \mathcal{E}_\Delta(s; \pi, z) = \frac{1}{K} \sum_{z' \in \mathcal{Z}} \mathbb{E}_{s^+ \sim V^{z'}(\cdot \mid s)} \left[ \frac{\mathcal{M}^{z'}(s^+ \mid s, \pi, z)}{V^{z'}(s^+ \mid s)} \log \frac{\mathcal{M}^{z'}(s^+ \mid s, \pi, z)}{\frac{1}{K} \sum_{z'' \in \mathcal{Z}} \mathcal{M}^{z''}(s^+ \mid s, \pi, z)} \right].
    \label{eq:e_delta_is}
\end{equation}

The complete policy objective is:
\begin{equation}
    \pi \leftarrow \arg\max_{\pi} \; \mathbb{E}_{(s,a) \sim \mathcal{D},\, z \sim p(z)}\Big[\mathcal{E}_\Delta\big(s; \pi, z\big) - \frac{\alpha_{\text{bc}}}{2} \big\|a - \mu_\theta(s,z)\big\|^2 \Big]
    \label{eq:policy_obj}
\end{equation}
where $\alpha_{\text{bc}}$ is a hyperparameter that enforces a behavioral constraint. The behavioral constraint is the log-likelihood of the dataset action under a unit-variance Gaussian centered at the policy mean. We train the Q, V networks and the policy jointly, taking a gradient step on their combined losses at every iteration (\cref{alg:offline-empowerment}).

\subsection{Estimating the Empowerment}
\label{sec:estimator}
Once training has converged, the learned policy $\pi$ approximately maximizes
the effective empowerment, so its effective empowerment $\tilde{\mathcal{E}}(s;\pi)$
serves as an estimate of the potential empowerment $\mathcal{E}(s)$. We compute
it by substituting the learned occupancy model $V^z_\theta$ into
\cref{eq:effective_emp} and approximating the expectation by sampling latents from the Gaussian parameterization of $V^z_\theta$ in \cref{eq:v_latent_param}:
\begin{equation}
    \hat{\mathcal{E}}(s) \;=\; \frac{1}{K} \sum_{z \in \mathcal{Z}} \frac{1}{M} \sum_{m=1}^{M}
    \left[ \log V^z_\theta(\psi_{z,m} \mid s) - \log \frac{1}{K} \sum_{z' \in \mathcal{Z}}
    V^{z'}_\theta(\psi_{z,m} \mid s) \right],
    \label{eq:emp_estimator}
\end{equation}
where $\psi_{z,m} \sim \mathcal{N}\big(\phi_V(s,z), \tfrac{d}{2} I\big)$ are $M$
latent samples per skill and $V^z_\theta(\psi \mid s)$ denotes
\cref{eq:v_latent_param} evaluated at latent $\psi$. This is the estimator we
evaluate in \cref{sec:experiments}.

\subsection{Additional Details}
\label{sec:additional-details}

\textbf{Action noise.} Every expectation over $a \sim \pi(\cdot \mid s, z)$, in $\mathcal{L}_V$ and in the policy objective (\cref{sec:policy-opt}), is estimated with reparameterized samples $a = \mathrm{clip}(\mu_\theta(s,z) + \eta,\, -1,\, 1)$.
Since the policy's actions are small perturbations of its mean, we also perturb the dataset actions with the same noise when training $Q_\theta$, $\tilde{a} = \mathrm{clip}(a + \eta,\, -1,\, 1)$, so that $Q_\theta$ is trained on the same kind of slightly noisy actions at which it is later evaluated.
\cref{tab:sigma-a} lists the value of $\sigma_a$ used for each dataset.

\textbf{Clipped Linear Exponential Loss.} Because $Q$ and $V$ operate in log-space, we regress in the exponentiated domain using a clipped linear-exponential loss for numerical stability, with the clipping threshold set to $t = 5$ to avoid exploding gradients; its exact form is given in \cref{sec:appendix-linexp}.

\begin{algorithm}[t]
\caption{Offline Empowerment Estimation}
\label{alg:offline-empowerment}
\begin{algorithmic}[1]
\Require Offline dataset $\mathcal{D}$ of transitions $(s, a, s')$; number of skills $K$; discount $\gamma$; behavioral constraint coefficient $\alpha_{\text{bc}}$; Polyak rate $\tau$
\State Initialize occupancy networks $Q_\theta, V_\theta$, target networks $Q_{\bar\theta} \gets Q_\theta$, $V_{\bar\theta} \gets V_\theta$, and policy $\pi(a \mid s, z)$
\For{each training step}
    \State Sample a batch of transitions $(s, a, s') \sim \mathcal{D}$
    \State Sample a random state $s^+ \sim \mathcal{D}$ and a skill $z \sim \mathrm{Unif}\{1, \dots, K\}$ for each transition
    \State Compute the occupancy losses $\mathcal{L}_Q = \mathcal{L}^{\mathrm{boot}}_Q + \mathcal{L}^{\mathrm{self}}_Q$ and $\mathcal{L}_V$ \Comment{Sec.~\ref{sec:occupancy}}
    \State Sample latent futures $\psi_{z',m} \sim \mathcal{N}\big(\phi_V(s, z'), \tfrac{d}{2} I\big)$ for each $z' \in \mathcal{Z}$, $m = 1, \dots, M$
    \State Sample actions $a_m \sim \pi(\cdot \mid s, z)$ with the reparameterization trick
    \State Estimate $\mathcal{E}_\Delta(s; \pi, z)$ with Eq.~\ref{eq:e_delta_is}, averaging over the samples \Comment{Sec.~\ref{sec:policy-opt}}
    \State Compute the policy loss $\mathcal{L}_\pi$ from Eq.~\ref{eq:policy_obj}
    \State Take one gradient step on $\mathcal{L}_Q + \mathcal{L}_V + \mathcal{L}_\pi$
    \State $\bar\theta \gets \tau \theta + (1 - \tau)\bar\theta$ \Comment{Polyak update of $Q_{\bar\theta}, V_{\bar\theta}$}
\EndFor
\State \Return $\hat{\mathcal{E}}(s)$ computed from $V_\theta$ with Eq.~\ref{eq:emp_estimator} \Comment{Sec.~\ref{sec:estimator}}
\end{algorithmic}
\end{algorithm}

\section{Experimental Setup}
\label{sec:approach}

Our experiments have two stages. We first train our empowerment estimator on offline OGBench datasets and compare it against offline adaptations of existing empowerment estimators (\cref{sec:offline-emp-plots,sec:baseline-emp-maps}). We then freeze the estimator and use it to guide online exploration on sparse-reward, goal-reaching tasks (\cref{sec:online-results}). Since computing the true empowerment over all policies is intractable in continuous environments, we evaluate the estimator qualitatively, in environments whose structure dictates where the empowerment should be high (e.g., toward the center of a maze rather than in its dead ends). Hyperparameters are given in \cref{tab:emp-hparams}.

\subsection{Baselines}
\label{sec:baselines}
Our method is the first to learn empowerment in continuous environments from offline data, so there are no direct baselines to compare to; instead, we adapt four existing online empowerment estimators to the offline setting. We give the core idea behind each baseline below; full derivations, architectures, and hyperparameters are given in \cref{sec:appendix-baselines}.

\textbf{Empowerment baselines.}
The baselines differ from each other, and from our method, along two orthogonal axes: \emph{which} mutual information they measure and \emph{how} they estimate it.
Three measure the \textit{action empowerment} $I(s^+;a\mid s)$, the mutual information between the action taken at $s$ in the dataset and the future state $s^+$ (\cref{sec:appendix-baselines}), and one measures the skill empowerment $I(s^+;z\mid s)$, as our method does; unlike our method, none of them computes it from learned occupancy measures.
\textbf{CF-Empowerment} (contrastive features) estimates $I(s^+;a\mid s)$ with a binary NCE contrastive critic~\citep{crl} using in-batch negatives.
\textbf{CF-FlowBC-Empowerment} replaces those in-batch negatives with actions from a learned flow-matching behavior-cloned policy~\citep{flow-matching}.
\textbf{DV-Empowerment} is architecturally identical but optimizes the Donsker--Varadhan bound~\citep{belghazi-mine} instead of InfoNCE.
\textbf{DADS-Empowerment} estimates the skill empowerment $I(s^+;z\mid s)$, matching our formalism, by jointly learning a discrete skill labeling and transition channel that maximize the Barber--Agakov bound~\citep{dads,barber-agakov}.

Online, errors in their learned models (e.g., dynamics or skill labelings) are corrected by fresh data, but offline, these methods suffer distribution shift and approximation errors that bias the estimate (\cref{sec:baseline-emp-maps}).

\subsection{Online Exploration with Offline Empowerment}
\label{sec:online-setup}

We also use the empowerment estimator $\hat{\mathcal{E}}(s)$ (\cref{eq:emp_estimator}), trained on $\mathcal{D}$ and then frozen, to guide exploration when an agent later solves online, sparse-reward, goal-reaching tasks in the same environment.

\textbf{Base agent.} Our online agents are built on contrastive RL (CRL)~\citep{crl}, a goal-conditioned actor-critic. The critic $f(s,a,g) = \phi(s,a)^\top \psi(g)$ is trained with a contrastive loss to tell whether $g$ is a future state of $(s,a)$ on the same trajectory, and the policy $\pi(a \mid s, g)$ maximizes $f$ along with a SAC-style auto-tuned entropy term~\citep{sac, sac-apps}. Every online agent is trained with Reinforcement Learning with Prior Data (RLPD)~\citep{rlpd}: each training batch is drawn half from the online replay buffer and half from $\mathcal{D}$.

\textbf{Empowerment bonus.} A naive way to use empowerment is to add it as an additional reward; however, this can have the undesired consequence of keeping the policy from leaving high-empowerment states. Instead, we would like the policy to visit high-empowerment states more often so it can explore around them, but not be penalized for leaving those high-empowerment states and continuing its trajectory into other parts of the state space. Consequently, we train a network $B_\omega(s,a)$ to predict the \textit{maximum} empowerment reached anywhere along the trajectory containing the transition $(s,a)$. For every complete trajectory $\tau = (s_0, a_0, \dots, s_T)$ in either the online or offline buffer, each of its transitions is regressed onto the highest empowerment reached along $\tau$:
\begin{equation}
\min_\omega \; \mathbb{E}_{\tau,\, t}\Big[\big(B_\omega(s_t, a_t) - \max_{0 \le k \le T} \hat{\mathcal{E}}(s_k)\big)^2\Big].
\label{eq:distill}
\end{equation}
Since the max is taken over both the future \textit{and} the history, leaving a high-empowerment state is not penalized, while actions that lead to one are rewarded. The policy then maximizes
\begin{equation}
\mathbb{E}_{(s,g),\; a \sim \pi(\cdot \mid s, g)}\big[f(s,a,g) + \beta\, B_\omega(s,a) - \alpha_{\text{ent}} \log \pi(a \mid s, g)\big],
\label{eq:online-actor}
\end{equation}
where $\beta$ is the bonus weight, and $\alpha_{\text{ent}}$ is the automatically tuned entropy temperature.

\textbf{Exploration baselines.} We compare against standard CRL and two exploration methods. \textbf{Random Network Distillation (RND)}~\citep{rnd} rewards novel states, measured by how poorly a trained predictor matches a fixed, randomly initialized network; this reward is learned into a separate critic $Q_{\text{RND}}(s,a)$, and the CRL policy maximizes $f(s,a,g) + \beta_{\text{RND}}\, Q_{\text{RND}}(s,a)$ along with the entropy term. \textbf{SUPE}~\citep{supe} pretrains a latent skill model on $\mathcal{D}$, then trains a high-level SAC policy online that reselects a skill every $H$ steps; offline data is relabeled with a skill and mixed into training with RLPD. We anneal both the RND and empowerment weight linearly to zero 50\% of the way through training, as we find this gives better results than holding constant.

\textbf{Combining empowerment with other exploration methods.} Because our bonus only adds a term to the policy objective, it can be combined with other exploration methods. We test the effect of combining our empowerment bonus with RND and also adding it to the SUPE high-level policy loss.

\textbf{Evaluation.} We evaluate on sparse-reward, goal-conditioned variants of OGBench environments in which the agent starts from a fixed state and must reach goals spread across the environment, as exploration is especially critical in these environments due to the sparsity of the reward signal: \textit{antmaze-medium} (an ant starts in a corner of the medium maze and must reach other cells of the maze), \textit{pointmaze-teleport} (a point must navigate a maze with random teleporters), \textit{antsoccer-arena} (the ant must push a ball to different goals), and \textit{cube-single} (a robot arm must move a cube to a different position on the ground).
\Cref{sec:appendix-online-envs} describes each task in detail.
Each task is paired with one or more OGBench datasets for $\mathcal{D}$, which differ in data quality: \textit{navigate} and \textit{play} contain long, goal-directed trajectories, \textit{stitch} contains only short trajectory segments that must be stitched together, \textit{noisy} adds heavy action noise, and \textit{explore} contains highly suboptimal, random-walk-like trajectories.
The empowerment estimator for each task is trained on the same dataset $\mathcal{D}$ used by the online agent for RLPD.
Implementation details and hyperparameters are given in \cref{sec:appendix-online}.

\section{Experiments}
\label{sec:experiments}

We answer two questions: (1) Does our method recover meaningful empowerment estimates from offline data? (2) Can we use the offline empowerment estimator to improve online exploration?

\subsection{Offline Empowerment Plots}
\label{sec:offline-emp-plots}

\begin{figure}[t]
    \centering
    \begin{subfigure}[t]{0.365\linewidth}
        \centering
        \includegraphics[width=\linewidth]{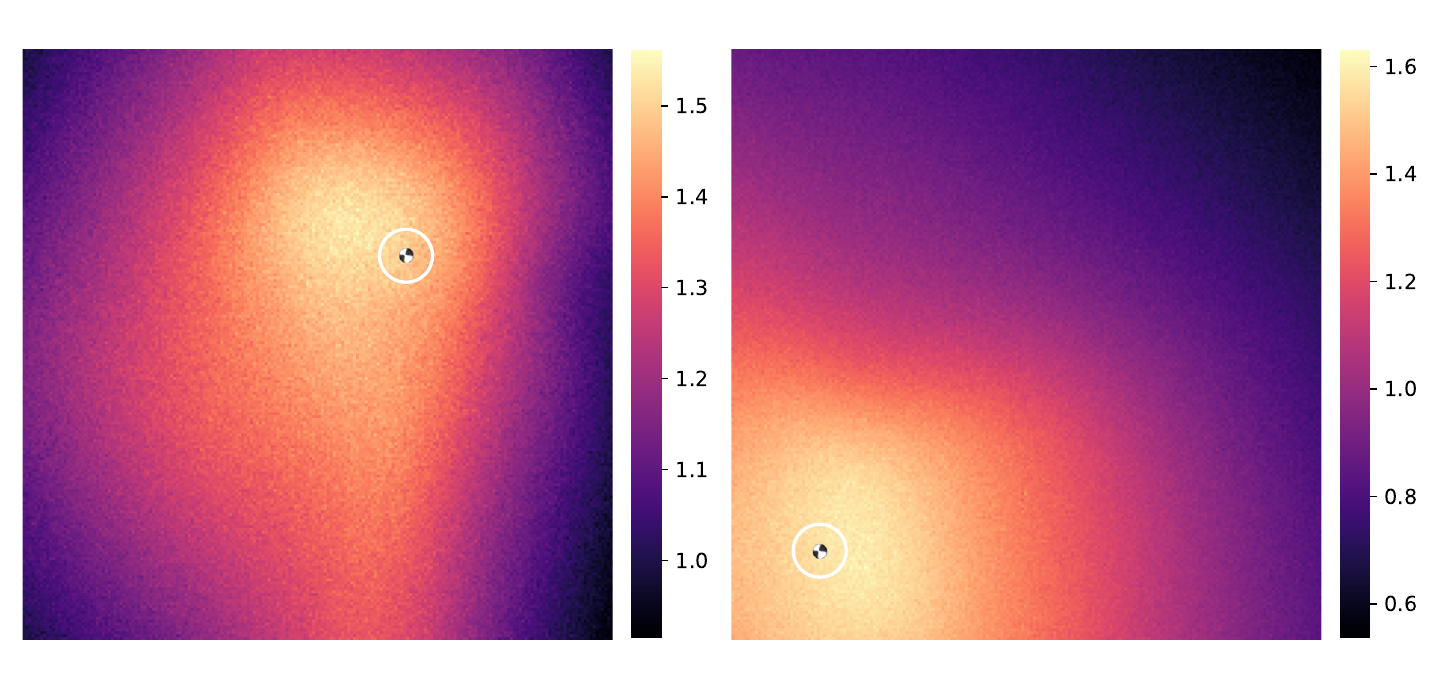}
        \caption{Ant-Soccer Arena.}
        \label{fig:emp-antsoccer-arena-navigate}
    \end{subfigure}
    \hspace{0.02\linewidth}
    \begin{subfigure}[t]{0.198\linewidth}
        \centering
        \includegraphics[width=\linewidth]{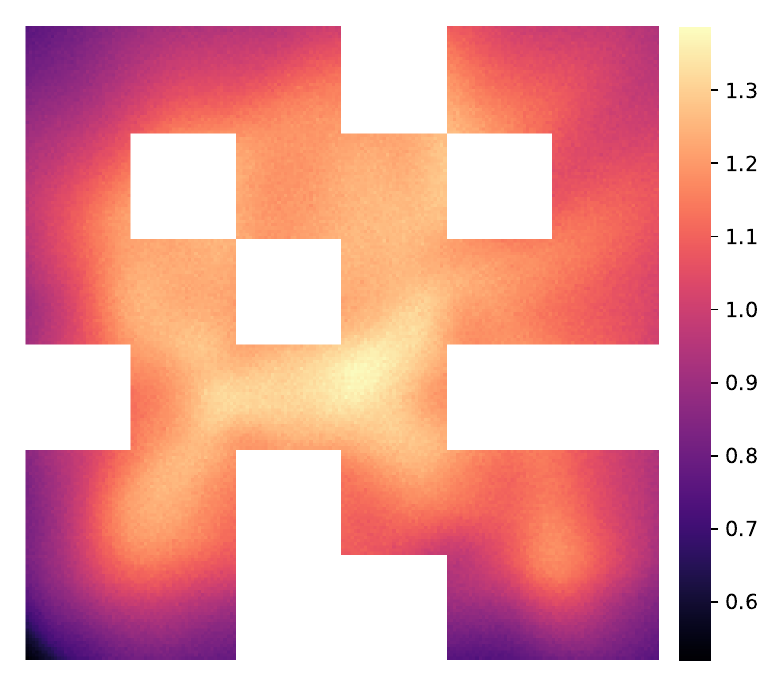}
        \caption{AntMaze.}
        \label{fig:emp-antmaze-medium-navigate}
    \end{subfigure}
    \hspace{0.02\linewidth}
    \begin{subfigure}[t]{0.262\linewidth}
        \centering
        \includegraphics[width=\linewidth]{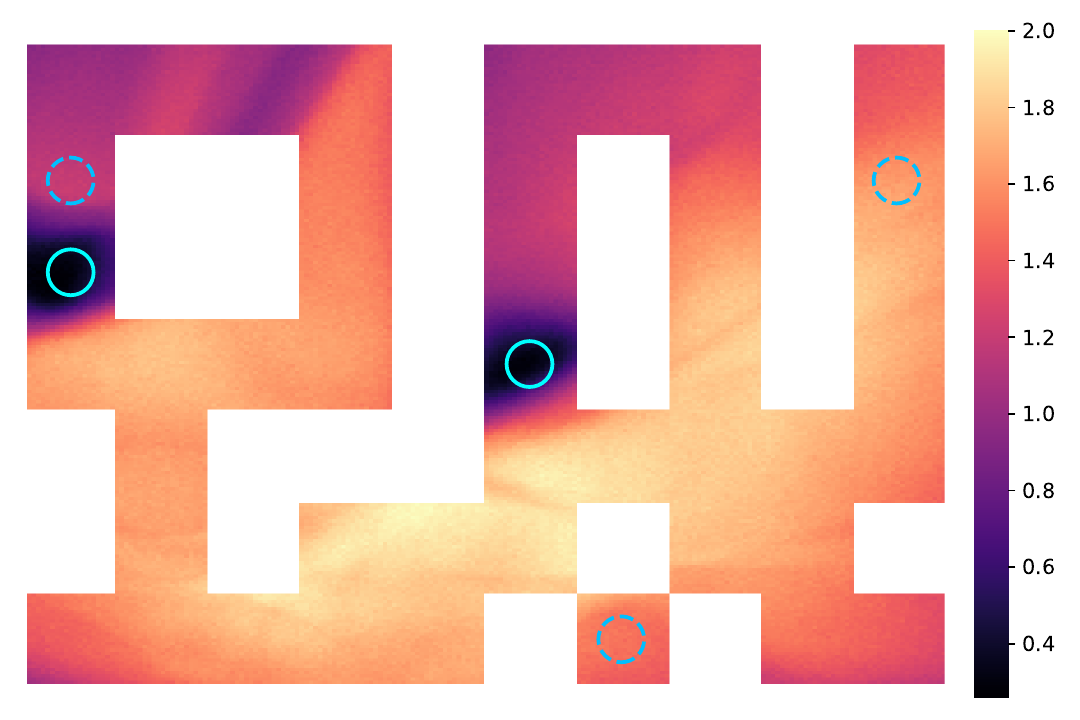}
        \caption{PointMaze-Teleport.}
        \label{fig:emp-pointmaze-teleport}
    \end{subfigure}

    \caption{Empowerment recovered by our offline estimator; axes are the agent's $(x,y)$ position. (a) \textit{antsoccer-arena-navigate-v0} with the ball (circled in white) fixed off-center (left) and cornered (right); each point's color is the empowerment with the ant there. (b) \textit{antmaze-medium-navigate-v0}. (c) \textit{pointmaze-teleport-navigate-v0}, with teleporter entrances (solid circles) and exits (dashed).}
    \label{fig:offline_heatmap}
\end{figure}

\cref{fig:offline_heatmap} shows the empowerment recovered by our offline estimator on three OGBench environments; in each case it matches what we intuitively expect.
On \textit{antsoccer-arena-navigate-v0} it peaks immediately around the ball in both plots,
where the ant can freely choose whether to approach, dribble, or move away from it.
On \textit{antmaze-medium-navigate-v0} it peaks at the maze's two corridor junctions and falls off in dead ends and corners.
On \textit{pointmaze-teleport-navigate-v0} it collapses at each one-way teleporter entrance, where the agent has no control over its random next state.

\begin{figure}[t]
    \centering
    \begin{subfigure}[t]{0.45\linewidth}
        \centering
        \includegraphics[width=\linewidth]{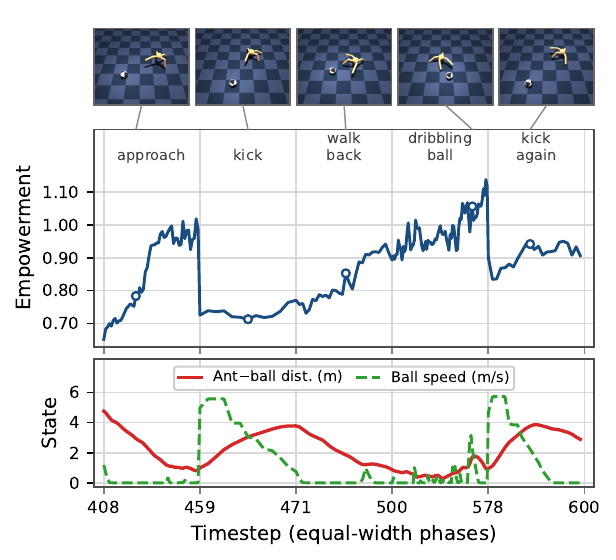}
        \caption{Ant-Soccer.}
        \label{fig:antsoccer-kick}
    \end{subfigure}
    \hfill
    \begin{subfigure}[t]{0.45\linewidth}
        \centering
        \includegraphics[width=\linewidth]{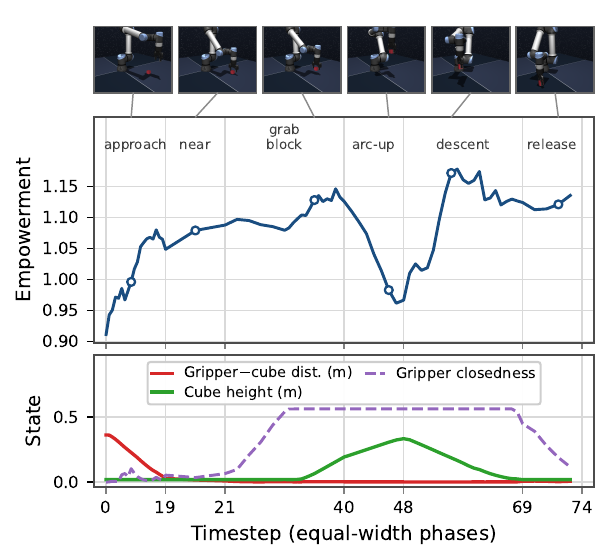}
        \caption{Cube.}
        \label{fig:cube-cycle}
    \end{subfigure}
    \caption{Empowerment (middle panel) recovered by our offline estimator along real dataset trajectories, alongside the underlying state (bottom). (a) \textit{antsoccer}: the ant walks up to the ball, kicks it away, walks back, dribbles it, and kicks it again. (b) \textit{cube-single}: one cycle in which the arm grabs the block, picks it up, and places it down in a new spot. Rendered frames (top) show the scene at the timestep marked by each hollow circle. The $x$-axis has timesteps at the phase boundaries as tick labels.}
    \label{fig:emp-trajectories}
\end{figure}

\cref{fig:emp-trajectories} shows the empowerment estimate along real dataset trajectories.
In \textit{antsoccer} (\cref{fig:antsoccer-kick}), empowerment rises as the ant approaches the ball, drops sharply the moment it kicks it, and stays low while the ball rolls out of reach, then climbs again as the ant walks back and dribbles, peaking just before the second kick with the ball at its feet. In \textit{cube} (\cref{fig:cube-cycle}), empowerment rises as the gripper approaches and grasps the block. Once the arm carries the block up into the air, the empowerment falls: this is because the gripper must stay closed to retain control of the block, giving the agent fewer degrees of freedom to control its future state. Empowerment rises as the arm descends and releases the block; here empowerment is high because, as in \textit{antsoccer}, the arm can choose to back off from the cube, or grab it and move it.

\subsection{Qualitative Comparison to Baselines}
\label{sec:baseline-emp-maps}

\begin{figure}[t]
    \centering
    \rotatebox{90}{\hspace{2.5pt}\scriptsize AntMaze}\hspace{2pt}%
    \includegraphics[width=0.66\linewidth, trim=0 34 0 0, clip]{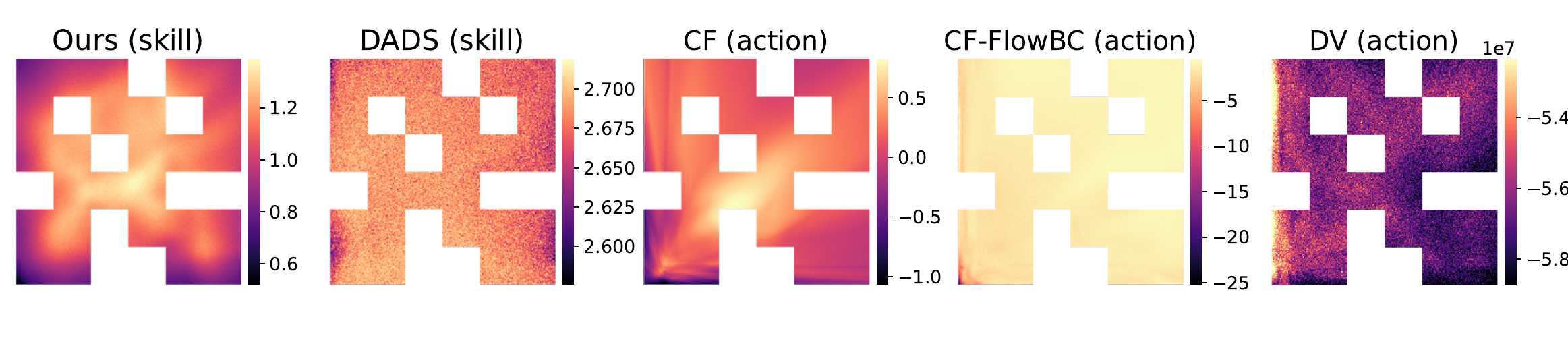}\\[2pt]
    \rotatebox{90}{\hspace{0.5pt}\scriptsize Ant-Soccer}\hspace{2pt}%
    \includegraphics[width=0.66\linewidth, trim=0 34 0 0, clip]{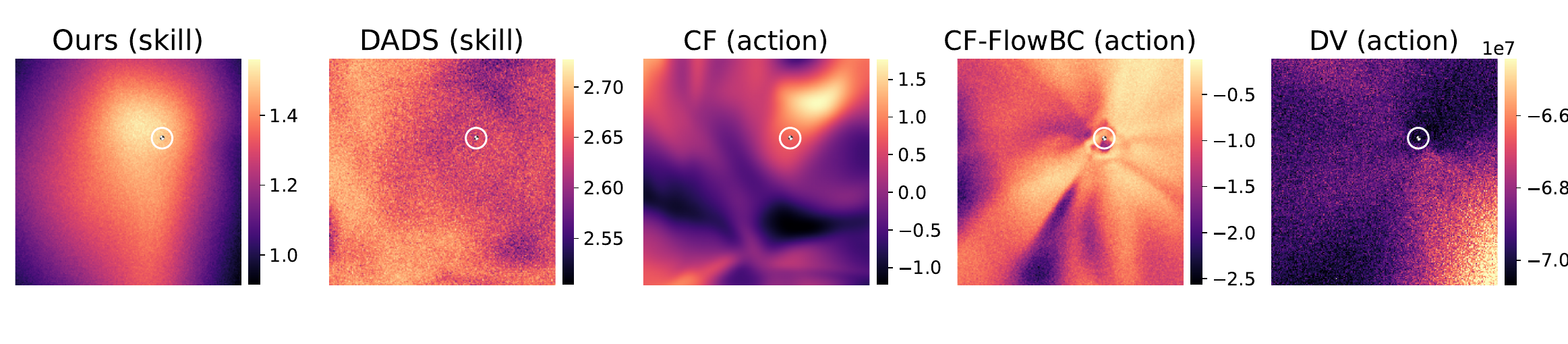}
    \caption{Empowerment maps for our method and the four baselines of \cref{sec:baselines}, on \textit{antmaze-medium-navigate} (top) and \textit{antsoccer-arena-navigate} with the ball (circled in white) fixed off-center (bottom). DADS is skill-empowerment; CF, CF-FlowBC, and DV are action-empowerment.}
    \label{fig:baseline-emp}
\end{figure}

\cref{fig:baseline-emp} shows $\hat{\mathcal{E}}(s)$ recovered by the four empowerment baselines of \cref{sec:baselines}. None recovers a comparable map.
\textbf{DADS-Empowerment} saturates at its ceiling of $\log K$ almost everywhere; \textbf{CF-Empowerment} learns a reasonable estimate, but is noisy and only captures the coarser trends; \textbf{CF-FlowBC-Empowerment} also fails. \textbf{DV-Empowerment}, whose bound is known to suffer from high variance~\citep{song2020understanding}, is virtually unbounded.

\begin{figure}[!t]
    \centering
    \begin{subfigure}[t]{\linewidth}
        \centering
        \includegraphics[width=0.8\linewidth]{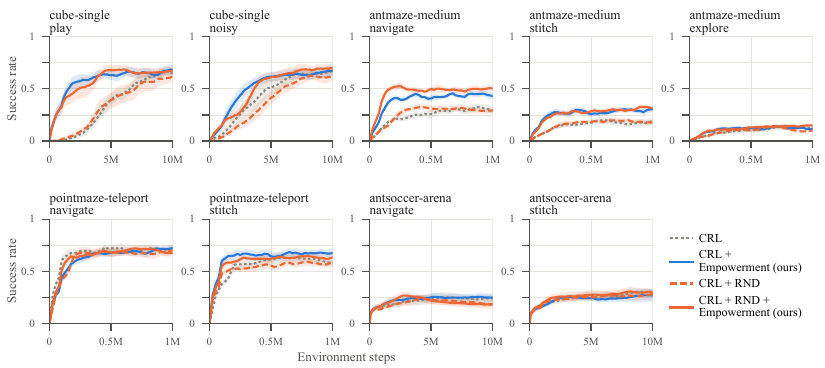}
        \caption{CRL and CRL + RND.}
        \label{fig:online-results-crl}
    \end{subfigure}
    \\[4pt]
    \begin{subfigure}[t]{\linewidth}
        \centering
        \includegraphics[width=0.8\linewidth]{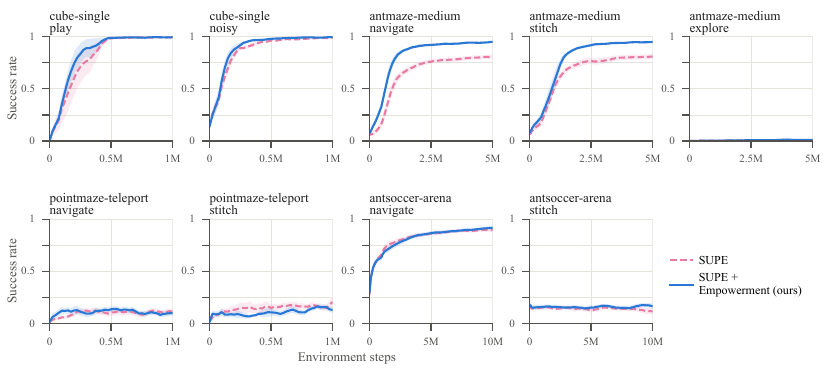}
        \caption{SUPE.}
        \label{fig:online-results-supe}
    \end{subfigure}
    \caption{Online success rate on each task (top line of each panel title) with each offline dataset (bottom line), for CRL, RND, and SUPE with and without our empowerment bonus. We use 5 seeds.}
    \label{fig:online-results}
\end{figure}

\subsection{Online Exploration}
\label{sec:online-results}

\cref{fig:online-results} compares our empowerment bonus against CRL, CRL + RND, and SUPE on every task and dataset (\cref{sec:online-setup}), and also shows the bonus combined with RND and SUPE. Across \textit{cube-single-play}, \textit{cube-single-noisy}, \textit{antmaze-medium-navigate}, \textit{antmaze-medium-stitch}, and \textit{pointmaze-teleport-stitch}, adding the empowerment bonus improves over baseline CRL, enabling faster learning and higher performance. On the other environments, our method still performs on par with CRL and does not hurt performance. Our method can also be combined with existing exploration methods. Adding the empowerment to CRL + RND leads to improved performance on \textit{cube-single}, \textit{antmaze-medium}, and \textit{pointmaze-teleport} over RND alone. Adding it to SUPE on \textit{cube} and \textit{antmaze} speeds up learning.

\textbf{Ablations.}
\label{sec:ablations}
\cref{sec:appendix-ablations} ablates the action noise scale $\sigma_a$ and behavioral constraint coefficient $\alpha_{\text{bc}}$ of our estimator, and the weight $\beta$ of the online empowerment bonus.

\FloatBarrier

\section{Conclusion}
\label{sec:conclusion}

In this work we introduced a method for estimating empowerment purely from offline data, by jointly learning an empowerment-maximizing policy together with its discounted state-occupancy measure.
Unlike prior methods, which rely on online rollouts of a skill-conditioned policy, our estimator trains on unlabeled offline trajectories and recovers an intuitive empowerment landscape across OGBench environments.
Our results show that the estimator can be used to improve online exploration on sparse, goal-conditioned RL tasks.
Beyond exploration, our estimator is general, and we hope future work will apply it to other problems and domains.

\bibliography{references}
\bibliographystyle{iclr2027_conference}

\appendix
\section{Appendix}
\label{sec:appendix}

\subsection{Hyperparameters}
\label{sec:appendix-hparams}

This section shows hyperparameters for our offline empowerment estimator (\cref{sec:background}). Hyperparameters for the four baselines are given alongside their algorithm descriptions in \cref{sec:appendix-baselines}.

\textbf{Offline Empowerment (Ours).} \cref{tab:emp-hparams} lists hyperparameters for the offline empowerment estimator and skill-conditioned policy of \cref{sec:background}.

\begin{table}[htbp]
    \centering
    \caption{Hyperparameters for offline empowerment pretraining (ours).}
    \label{tab:emp-hparams}
    \begin{tabular}{@{}ll@{}}
        \toprule
        Hyperparameter                                         & Value            \\
        \midrule
        Number of skills $K$                                   & 50               \\
        $Q$/$V$ network width / depth                          & 512 / 3          \\
        Policy network width / depth                           & 512 / 3          \\
        Value latent dimension $d$                             & 256              \\
        Layer normalization                                    & yes              \\
        Learning rate ($Q$/$V$ and policy)                     & $3\times10^{-4}$ \\
        Batch size                                             & 1024             \\
        Discount $\gamma$                                      & 0.99             \\
        Target network update rate $\tau$                      & 0.005            \\
        Clipped linexp threshold $t$                           & 5                \\
        Action noise std $\sigma_a$                            & 0.01             \\
        Behavioral constraint coefficient $\alpha_{\text{bc}}$ & 0.001            \\
        Future-state samples per update                        & 1                \\
        Training steps                                         & $10^6$           \\
        \bottomrule
    \end{tabular}
\end{table}

\begin{table}[htbp]
    \centering
    \caption{Action noise std $\sigma_a$ of the empowerment estimator trained on each offline dataset.}
    \label{tab:sigma-a}
    \small
    \begin{tabular}{@{}llc@{}}
        \toprule
        Environment        & Dataset  & $\sigma_a$ \\
        \midrule
        antmaze-medium     & navigate & 0.01 \\
        antmaze-medium     & stitch   & 0.01 \\
        antmaze-medium     & explore  & 0.01 \\
        antsoccer-arena    & navigate & 0.01 \\
        antsoccer-arena    & stitch   & 0.01 \\
        pointmaze-teleport & navigate & 0.01 \\
        pointmaze-teleport & stitch   & 0.01 \\
        cube-single        & play     & 0.01 \\
        cube-single        & noisy    & 0.01 \\
        \bottomrule
    \end{tabular}
\end{table}

\FloatBarrier

\subsection{Ablations}
\label{sec:appendix-ablations}

This section ablates three hyperparameters of our method: the policy's action noise scale $\sigma_a$ (\cref{sec:prelims,sec:additional-details}), the behavioral constraint coefficient $\alpha_{\text{bc}}$ (\cref{sec:policy-opt}), and the weight $\beta$ of the online empowerment bonus (\cref{sec:online-setup}). For $\sigma_a$ and $\alpha_{\text{bc}}$, we sweep the hyperparameter, holding everything else at the values in \cref{tab:emp-hparams}, and track the validation empowerment estimate during training.

\subsubsection{Action Noise Scale}
\label{sec:appendix-ablation-noise}

\begin{figure}[htbp]
    \centering
    \includegraphics[width=0.55\linewidth]{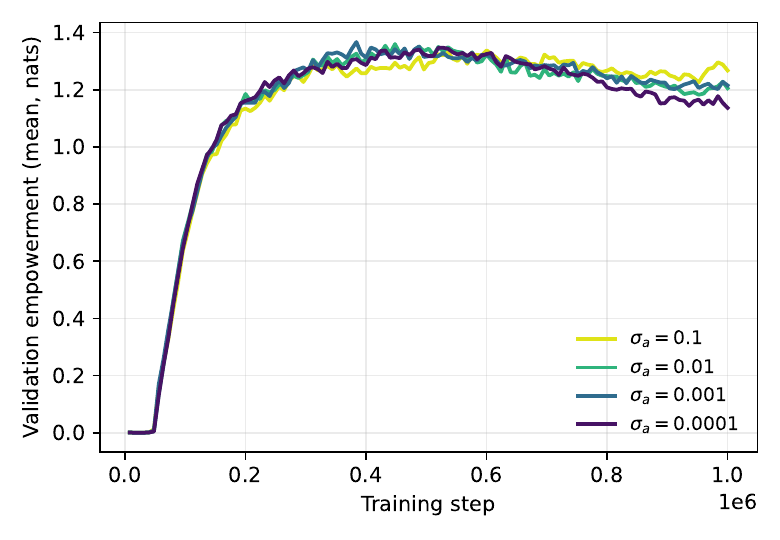}
    \caption{Validation empowerment over training for four action noise scales $\sigma_a$, on \textit{antmaze-medium-navigate-v0}. The converged empowerment is robust to $\sigma_a$; all other hyperparameters are fixed at the values in \cref{tab:emp-hparams}.}
    \label{fig:noise-sweep}
\end{figure}

\cref{fig:noise-sweep} sweeps $\sigma_a$ over three orders of magnitude, $\sigma_a \in \{10^{-4}, 10^{-3}, 10^{-2}, 10^{-1}\}$, on \textit{antmaze-medium-navigate-v0}. The converged empowerment estimate is highly robust to the action noise scale.

\subsubsection{Behavioral Constraint Coefficient}
\label{sec:appendix-ablation-bc}

\begin{figure}[htbp]
    \centering
    \includegraphics[width=0.55\linewidth]{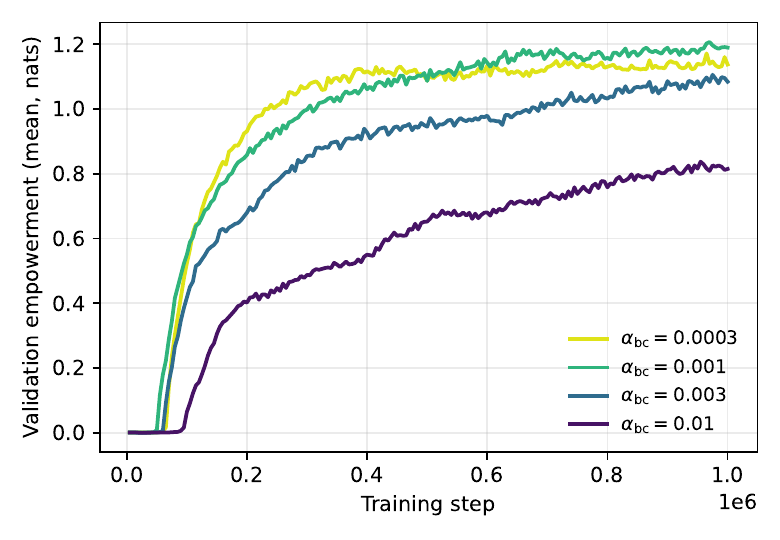}
    \caption{Validation empowerment over training for four behavioral constraint coefficients $\alpha_{\text{bc}}$, on \textit{antsoccer-arena-stitch-v0}. Empowerment is roughly flat until $\alpha_{\text{bc}}=10^{-2}$, where it collapses.}
    \label{fig:bc-sweep}
\end{figure}

\cref{fig:bc-sweep} sweeps the behavioral constraint coefficient $\alpha_{\text{bc}} \in \{3\times10^{-4}, 10^{-3}, 3\times10^{-3}, 10^{-2}\}$ on \textit{antsoccer-arena-stitch-v0}. Because $\alpha_{\text{bc}}$ directly weights how strongly the policy is pulled toward the offline behavior policy, a value that is too large prevents the policy from ever departing from the dataset's own behavior far enough to realize the higher-empowerment behaviors. Conversely, $\alpha_{\text{bc}}$ cannot be too low as it must regularize the skill-conditioned policy to the offline data.

\begin{figure}[htbp]
    \centering
    \includegraphics[width=0.85\linewidth]{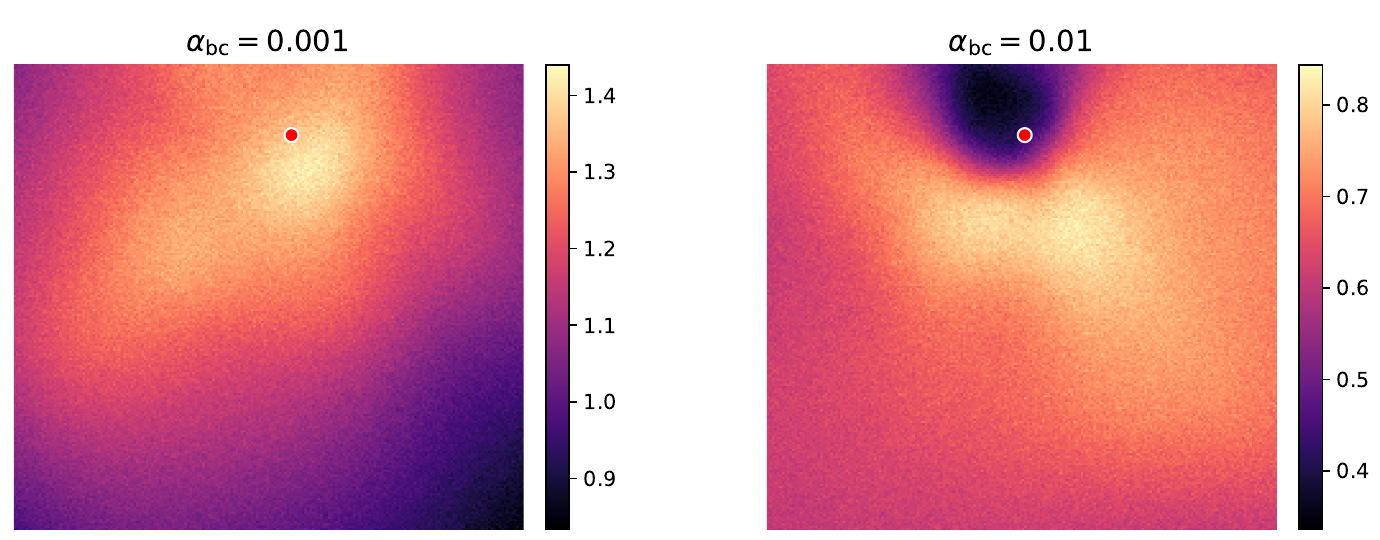}
    \caption{Empowerment maps on \textit{antsoccer-arena-stitch-v0} at matched ball position, for $\alpha_{\text{bc}}=10^{-3}$ (left) vs.\ $\alpha_{\text{bc}}=10^{-2}$ (right). Heavy BC regularization produces a hole in empowerment exactly at the ball, where the offline data is dominated by a single (dribbling) behavior.}
    \label{fig:bc-map-ablation}
\end{figure}

\textbf{Qualitative failure mode.} \cref{fig:bc-map-ablation} shows the empowerment map for two checkpoints on \textit{antsoccer-arena-stitch-v0} that differ only in $\alpha_{\text{bc}}$. At $\alpha_{\text{bc}}=10^{-3}$, empowerment peaks tightly around the ball, matching the desired pattern in \cref{fig:emp-antsoccer-arena-navigate}. At $\alpha_{\text{bc}}=10^{-2}$, the map instead shows a distinctive \emph{hole} exactly at the ball. This is the collapse from \cref{fig:bc-sweep} made spatially concrete: on the stitch dataset, the offline trajectories that pass near the ball are dominated by a single behavior---dribbling it---so a heavily BC-regularized policy is pulled toward dribbling near the ball regardless of skill, and never discovers the alternative high-empowerment behaviors (e.g. deliberately moving away from the ball) that would make the skill distinguishable. Notably, this failure mode is specific to the stitch dataset; since the \textit{navigate} dataset contains trajectories of the ant doing many behaviors near the ball, a much wider range of BC values works well. On low-coverage data such as \textit{stitch}, $\alpha_{\text{bc}}$ must be kept low enough so that the policy can generalize beyond the offline data and learn empowerment-maximizing skills. In general, we find that there is around a single order of magnitude range in which the BC coefficient should lie to give good empowerment estimates, for all environments.

\subsubsection{Empowerment Bonus Weight}
\label{sec:appendix-ablation-beta}

\begin{figure}[htbp]
    \centering
    \includegraphics[width=0.69\linewidth]{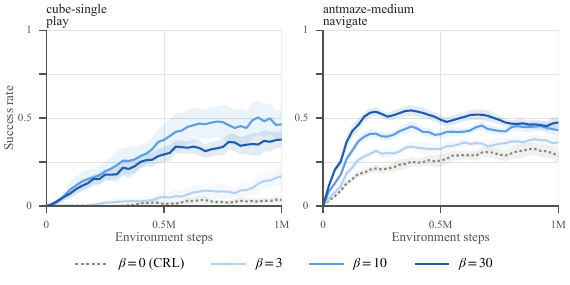}
    \caption{Online success rate of CRL + Empowerment for different weights $\beta$ of the empowerment bonus, on each task (top line of each panel title) with each offline dataset (bottom line); $\beta = 0$ corresponds to CRL. Lines show the mean over 5 seeds and shaded regions show $\pm 1$ standard error of the mean, smoothed as in \cref{fig:online-results}.}
    \label{fig:bonus-ablation}
\end{figure}

\cref{fig:bonus-ablation} varies the weight $\beta$ of the empowerment bonus in the policy objective of \cref{eq:online-actor}, comparing with CRL ($\beta = 0$). We sweep $\beta \in \{3, 10, 30\}$ on \textit{cube-single} (\textit{play}) and \textit{antmaze-medium} (\textit{navigate}). On \textit{cube-single}, performance is non-monotonic in $\beta$: $\beta = 10$ performs best, while $\beta = 3$ improves only slightly over CRL. On \textit{antmaze-medium}, every $\beta$ improves over CRL. In general, like other exploration methods, $\beta$ must be chosen to encourage the policy to visit high-empowerment states without dominating the policy objective.

\FloatBarrier

\subsection{Online Exploration Details}
\label{sec:appendix-online}

This section gives implementation details for the online experiments of \cref{sec:online-setup}.

\subsubsection{Online Environments}
\label{sec:appendix-online-envs}

All online tasks are goal-conditioned variants of OGBench environments~\citep{ogbench} with a single fixed start state and a fixed set of goals. At the start of every episode, one goal is drawn uniformly from the set and given to the agent, and neither the start nor the goal is perturbed by noise. The reward is sparse: the agent receives no signal until it reaches the goal, and an evaluation episode counts as a success if the agent reaches the goal within the episode horizon. \cref{tab:online-envs} summarizes the tasks, and the offline datasets used for RLPD and for pretraining the empowerment estimator. In \cref{fig:online-results}, shaded regions show $\pm 1$ standard error of the mean over seeds.

\begin{itemize}
    \item \textbf{antmaze-medium:} The ant starts in the bottom-left corner of OGBench's medium maze, and the goal is the center of any of the 25 other free cells of the maze, so most goals require navigating several corridors and turns.
    \item \textbf{pointmaze-teleport:} A point mass starts in the bottom-left corner of the teleport maze, and the goal is the center of any of 41 other free cells. The maze contains teleporters that send the agent to a random exit; we exclude the two teleporter entrances, where the agent cannot remain inside the goal region, and one walled-off pocket that can only be entered by a lucky teleport and cannot be left.
    \item \textbf{antsoccer-arena:} The ant starts in the open arena about three units behind the ball, so it must first walk to the ball and then push it to one of eight goals, each one to three units beyond the ball's start on the side away from the ant. The episode horizon is 500 steps.
    \item \textbf{cube-single:} A single cube starts at a fixed position on the table, and the goal is one of 30 positions on a $5 \times 6$ grid covering the table. The closest goals are just outside the success radius of the start, so every task requires picking up or pushing the cube. The episode horizon is 200 steps.
\end{itemize}

\begin{table}[htbp]
    \centering
    \caption{Online tasks and the offline datasets $\mathcal{D}$ paired with them.}
    \label{tab:online-envs}
    \small
    \begin{tabular}{@{}llll@{}}
        \toprule
        Task               & Goals & Horizon & Offline datasets          \\
        \midrule
        antmaze-medium     & 25    & 1000    & navigate, stitch, explore \\
        pointmaze-teleport & 41    & 1000    & navigate, stitch          \\
        antsoccer-arena    & 8     & 500     & navigate, stitch          \\
        cube-single        & 30    & 200     & play, noisy               \\
        \bottomrule
    \end{tabular}
\end{table}

\subsubsection{Agents}

\textbf{Training loop.} All online agents interact with a single environment. After an initial warm-up period, the agent collects a fixed number of environment steps and then performs a fixed number of gradient updates per environment step, on batches drawn from the online replay buffer and the offline dataset $\mathcal{D}$ in equal proportion (RLPD). For both sources, CRL goals are future states of the same trajectory, sampled at a geometrically distributed offset.

\textbf{Empowerment bonus.} The estimator $\hat{\mathcal{E}}$ is the pretrained, frozen model of \cref{sec:estimator}, trained on the same dataset $\mathcal{D}$ that RLPD uses. We evaluate $\hat{\mathcal{E}}(s)$ once for every state of $\mathcal{D}$ before training, and once for every new online state in a single batched call before each round of updates. The regression target in \cref{eq:distill} is shifted by the mean of $\hat{\mathcal{E}}$ over $\mathcal{D}$; this shifts $B_\omega$ by a constant and leaves the policy gradient unchanged, but keeps the regression well conditioned. $B_\omega$ has two heads, both trained with \cref{eq:distill}, and the policy uses their minimum. Transitions of an online episode that has not yet ended are excluded from the regression until the episode ends and its target is known; they are used by the CRL critic and the policy as usual.

\textbf{RND.} The target and predictor networks share one architecture and take the next state $s'$ as input, normalized by a running mean and standard deviation of online states and clipped to $[-5, 5]$. The intrinsic reward is the squared error between the predictor and target outputs, divided by a running standard deviation of the discounted intrinsic return. $Q_{\text{RND}}$ is trained with a twin-critic Bellman backup~\citep{td3} that does not stop at episode ends, as in~\citet{rnd}. The predictor, the running statistics, $Q_{\text{RND}}$, and the policy's RND bonus all use online transitions only.

\textbf{CRL + RND + Empowerment.} The two bonuses are trained independently exactly as above, with their own weights $\beta$ and $\beta_{\text{RND}}$.

\textbf{SUPE.} We follow the implementation of~\citet{supe}. The skill model is an OPAL-style VAE~\citep{opal} with a continuous latent, trained on $\mathcal{D}$ over segments of $H = 4$ steps (8 for soccer) and then frozen. The high-level policy is a tanh-Gaussian SAC actor over the latent, with an ensemble of 10 layer-normalized~\citep{layer-norm} critics. Each offline segment inside a single trajectory is labeled with the posterior mean of its latent. Following SUPE's treatment of sparse $-1/0$ tasks, at every update its reward is set to the dataset minimum plus an RND bonus, and its termination is set to the prediction of a reward model trained on online data.

\textbf{SUPE + Empowerment.} $B_\omega(s,u)$ has the same architecture as for CRL, takes the high-level state and skill $u$ as input, and is regressed onto the same targets as in \cref{eq:distill}. The high-level actor loss becomes $\alpha_{\text{ent}} \log \pi(u \mid s, g) - \bar{Q}(s,u,g) - \beta\, B_\omega(s,u)$, where $\bar{Q}$ is the mean over the critic ensemble.

\textbf{Hyperparameter selection.} For each task and dataset, we perform a sweep over the weight of each exploration bonus and report the best-performing setting: $\beta \in \{3, 10, 30\}$ for the empowerment bonus and $\beta_{\text{RND}} \in \{0.3, 1, 3\}$ for RND. For SUPE + Empowerment, we sweep $\beta \in \{10, 30\}$. For CRL + RND + Empowerment, we evaluate two configurations: the best weights from the individual sweeps and both weights halved. \cref{tab:online-bonus-weights} lists the weights used for each task; all other hyperparameters are shared across tasks (\cref{tab:online-shared-hparams,tab:online-bonus-hparams}).

\begin{table}[htbp]
    \centering
    \caption{Hyperparameters shared by all online agents (CRL + RLPD).}
    \label{tab:online-shared-hparams}
    \begin{tabular}{@{}ll@{}}
        \toprule
        Hyperparameter                        & Value \\
        \midrule
        Actor network width / depth & 512 / 3 \\
        Critic network width / depth & 512 / 3 \\
        Critic latent dimension & 512 \\
        Layer normalization & yes \\
        Learning rate & $3 \times 10^{-4}$ \\
        Batch size & 1024 \\
        Offline fraction of each batch (RLPD) & 0.5 \\
        Discount $\gamma$ & 0.99 \\
        Target entropy & $-\frac{1}{2}\dim(\mathcal{A})$ \\
        Environment steps per update round & 50 \\
        Updates per environment step & 1 \\
        Warm-up environment steps & 1000 \\
        Replay buffer size & $10^6$ \\
        \bottomrule
    \end{tabular}
\end{table}

\begin{table}[htbp]
    \centering
    \caption{Hyperparameters of the exploration bonuses.}
    \label{tab:online-bonus-hparams}
    \begin{tabular}{@{}ll@{}}
        \toprule
        Hyperparameter                                         & Value \\
        \midrule
        \multicolumn{2}{@{}l}{\textit{Empowerment bonus (ours)}} \\
        $B_\omega$ network width / depth & 512 / 3 (two heads) \\
        Bonus weight $\beta$ & per task (\cref{tab:online-bonus-weights}) \\
        Future-state samples per skill for $\hat{\mathcal{E}}$ & 64 \\
        \midrule
        \multicolumn{2}{@{}l}{\textit{RND}} \\
        Target / predictor network width / depth & 512 / 2 \\
        Embedding dimension & 512 \\
        Predictor learning rate & $10^{-4}$ \\
        $Q_{\text{RND}}$ network width / depth & 512 / 3 (two heads) \\
        $Q_{\text{RND}}$ discount & 0.99 \\
        $Q_{\text{RND}}$ target update rate $\tau$ & 0.005 \\
        Bonus weight $\beta_{\text{RND}}$ & per task (\cref{tab:online-bonus-weights}) \\
        \midrule
        \multicolumn{2}{@{}l}{\textit{SUPE}} \\
        Skill horizon $H$ & 4 (8 on antsoccer-arena) \\
        Skill latent dimension & 8 \\
        OPAL KL coefficient & 0.1 (0.2 on cube-single) \\
        Number of critics / critics in target minimum & 10 / 1 (10 / 2 on cube-single) \\
        Discount & 0.995 \\
        RND coefficient & 8 \\
        Warm-up environment steps & 5000 \\
        \bottomrule
    \end{tabular}
\end{table}

\begin{table}[htbp]
    \centering
    \caption{Bonus weights used for each task and dataset in \cref{fig:online-results}.}
    \label{tab:online-bonus-weights}
    \small
    \begin{tabular}{@{}llcccc@{}}
        \toprule
        Task & Dataset & CRL + Emp. & CRL + RND & CRL + RND + Emp. & SUPE + Emp. \\
         & & $\beta$ & $\beta_{\text{RND}}$ & $\beta$ / $\beta_{\text{RND}}$ & $\beta$ \\
        \midrule
        cube-single & play & 10 & 1 & 10 / 1 & 30 \\
        cube-single & noisy & 30 & 1 & 30 / 1 & 10 \\
        antmaze-medium & navigate & 30 & 3 & 30 / 3 & 10 \\
        antmaze-medium & stitch & 10 & 3 & 10 / 3 & 10 \\
        antmaze-medium & explore & 10 & 3 & 10 / 3 & 10 \\
        pointmaze-teleport & navigate & 30 & 1 & 15 / 0.5 & 30 \\
        pointmaze-teleport & stitch & 10 & 0.3 & 10 / 0.3 & 30 \\
        antsoccer-arena & navigate & 10 & 1 & 10 / 1 & 10 \\
        antsoccer-arena & stitch & 10 & 1 & 10 / 1 & 10 \\
        \bottomrule
    \end{tabular}
\end{table}

\FloatBarrier

\subsection{Compute Resources}
\label{sec:appendix-compute}

All offline pretraining runs were trained on an internal cluster of NVIDIA A5000 and A6000 GPUs, with a single GPU per run.

\subsection{Clipped Linear Exponential Loss}
\label{sec:appendix-linexp}

Because $Q$ and $V$ operate in log-space (\cref{sec:background}), we regress in the exponentiated domain for numerical stability. For a target $y$ and prediction $\hat{y}$, we regress $e^{-\hat{y}}$ to $\kappa e^{-y}$ (for a target scale $\kappa$) using
\begin{equation}
    \ell_{\mathrm{linexp}}(\hat{y}, y;\, \kappa) \;=\;
    \begin{cases}
        \kappa\, e^{\hat{y} - y} \;-\; \hat{y},                                               & \hat{y} - y < t,    \\[6pt]
        \big(\kappa e^{t} - 1\big)\big(\hat{y} - y - t\big) \;+\; \kappa e^{t} \;-\; (y + t), & \hat{y} - y \geq t.
    \end{cases}
    \label{eq:clipped-linexp}
\end{equation}
The second branch is simply a linearization of the loss about the point $\hat{y} = y + t$, applied whenever $\hat{y} - y$ exceeds the threshold $t = 5$, which prevents the exponential term from producing exploding gradients.

\subsection{Baseline Algorithm Details}
\label{sec:appendix-baselines}

This section gives full derivations, network architectures, and hyperparameters for the four baselines summarized in \cref{sec:baselines}.
All four baselines are trained on triples $(s,a,s^+) \sim \mathcal{D}$, where $s^+$ is a future state of $s$ on the same dataset trajectory, drawn at a geometrically distributed offset $k \geq 1$, with probability $(1-\gamma)\gamma^{k-1}$. This differs from our method, which draws $s^+$ uniformly from $\mathcal{D}$ and recovers the dependence on $(s,a)$ through TD backups.
With the exception of DADS-Empowerment (\cref{sec:appendix-dads}), the baselines estimate the \emph{action empowerment}
\begin{equation}
I(s^+; a \mid s) \triangleq \mathbb{E}_{a \sim \pi_\beta(\cdot \mid s),\, s^+ \sim p(\cdot \mid s, a)}\!\left[ \log \frac{p(s^+ \mid s, a)}{p(s^+ \mid s)} \right],
\end{equation}
that is, the mutual information between the primitive action $a \sim \pi_\beta(\cdot\mid s)$ actually taken at $s$ under the behavior policy $\pi_\beta$ and the resulting future state $s^+$, in place of the mutual information between a skill $z$ and $s^+$ used by our method.

\subsubsection{CF-Empowerment}
\label{sec:appendix-crl}

\textbf{Overview.} CF-Empowerment estimates the action empowerment $I(s^+;a\mid s)$ by first fitting two contrastive critics that share a future-state encoder, then distilling the resulting pointwise mutual-information estimator into a small feedforward network that can be evaluated at any state without re-running the contrastive critics.

\textbf{Architecture.} Three MLP trunks map into a shared latent space of dimension $d$: an action-conditioned encoder $\phi(s,a)$, a state-only encoder $\phi'(s)$, and a future-state encoder $\psi(s^+)$, all mapping into $\mathbb{R}^d$. A separate empowerment network $E_\theta(s) \in \mathbb{R}$ is a feedforward MLP that only depends on the current state. There is no policy network.

\textbf{Training objective.} Given a minibatch of $B$ triples $\{(s_i,a_i,s^+_i)\}_{i=1}^B$, define the bilinear logits
\begin{equation}
    \ell^a_{ij} = \frac{\phi(s_i,a_i)^\top \psi(s^+_j)}{\sqrt{d}}, \qquad \ell^s_{ij} = \frac{\phi'(s_i)^\top \psi(s^+_j)}{\sqrt{d}}.
\end{equation}
The critics are trained with a binary NCE objective~\citep{crl}, treating $s^+_i$ as the positive for $(s_i,a_i)$ (resp.\ $s_i$) and every other $s^+_j$ in the batch as a negative:
\begin{equation}
    \mathcal{L}_{\text{contrastive}} = \sum_{i,j=1}^B \Big[-\mathbf{1}\{i=j\}\log\sigma(\ell^a_{ij}) - \mathbf{1}\{i\neq j\}\log\big(1-\sigma(\ell^a_{ij})\big)\Big] + \Big[\text{same expression with } \ell^s_{ij}\Big].
\end{equation}
At the optimum, $\ell^a(s,a,s^+) \approx \log p(s^+\mid s,a) - \log p(s^+) + c$ and $\ell^s(s,s^+) \approx \log p(s^+\mid s) - \log p(s^+) + c$ for the same constant $c$, so $T(s,a,s^+) \triangleq \ell^a(s,a,s^+) - \ell^s(s,s^+)$ estimates the conditional mutual-information integrand $\log p(s^+\mid s,a) - \log p(s^+\mid s)$. After a fixed number of pretraining steps, the critics' parameters are frozen and the empowerment network is trained by regression onto $T$:
\begin{equation}
    \mathcal{L}_{\text{distill}} = \E_{(s,a,s^+)\sim\mathcal{D}}\Big[\big(E_\theta(s) - T(s,a,s^+)\big)^2\Big].
\end{equation}
At convergence $E_\theta(s) \to \E_{(a,s^+)\sim p(a,s^+\mid s)}[T(s,a,s^+)] = I(s^+;a\mid s)$, i.e., the network converges to an estimate of the action empowerment at $s$.

\textbf{Offline adaptation.} In the online setting, the negatives for the contrastive critics would be drawn by rolling out many actions from the same state; here they are instead the other $(s_j,a_j,s^+_j)$ triples already present in the offline minibatch (in-batch negatives), and $a$ ranges only over the fixed offline behavior policy rather than an arbitrary exploration policy.

\textbf{Hyperparameters.} Learning rate $10^{-3}$; batch size $512$; critic hidden dims $(256,256)$ with layer normalization; latent dimension $d=256$; empowerment network hidden dims $(256,256)$; contrastive pretraining for $5\times 10^5$ steps before switching to distillation.

\subsubsection{CF-FlowBC-Empowerment}
\label{sec:appendix-crl-flowbc}

\textbf{Overview.} CF-FlowBC-Empowerment also estimates the action empowerment $I(s^+;a\mid s)$, but replaces the in-batch negatives of CF-Empowerment with negatives drawn from a learned behavior-cloned policy, and replaces the bilinear critic with a single scalar critic $T(s^+,a,s)$ trained with an explicit InfoNCE lower bound.

\textbf{Architecture.} Two flow-matching models~\citep{flow-matching} are trained on the offline data: a behavior-cloned policy $\pi_{\text{bc}}(a\mid s)$ and a dynamics flow $f_{\text{dyn}}(s^+\mid s,a)$ modeling the discounted future state. Both share the same velocity-field MLP architecture $v_\theta(x_t,t,\text{cond})$, conditioned on $s$ (for $\pi_{\text{bc}}$) or on $(s,a)$ (for $f_{\text{dyn}}$); samples are drawn by Euler-integrating the learned velocity field for $8$ steps from Gaussian noise, with sampled actions clipped to the action bounds. A separate scalar critic $T(s^+,a,s)$, an MLP over the concatenation of its three inputs, provides the mutual-information estimate.

\textbf{Training objective.} Both flow models are trained with the standard conditional flow-matching loss,
\begin{equation}
    \mathcal{L}_{\text{flow}}(\text{cond},x) = \E_{x_0\sim\mathcal{N}(0,I),\, t\sim\text{Unif}(0,1)}\Big[\big\|v_\theta\big((1-t)x_0+tx,\, t,\, \text{cond}\big) - (x-x_0)\big\|^2\Big],
\end{equation}
instantiated with $(\text{cond},x)=(s,a)$ for $\pi_{\text{bc}}$ and $(\text{cond},x)=((s,a),s^+)$ for $f_{\text{dyn}}$. Given a positive triple $(s,a,s^+)$, $N$ negative actions $a_1,\dots,a_N \overset{\text{iid}}{\sim} \pi_{\text{bc}}(\cdot\mid s)$ are sampled and the critic is trained to maximize the categorical InfoNCE bound
\begin{align}
    b(s,a,s^+)                  & = \log(N{+}1) + \log \frac{e^{T(s^+,a,s)}}{e^{T(s^+,a,s)} + \sum_{i=1}^N e^{T(s^+,a_i,s)}}, \\
    \mathcal{L}_{\text{critic}} & = \log(N{+}1) - \E\big[b(s,a,s^+)\big].
\end{align}
The reported estimate $\hat{I}(s^+;a\mid s) = \log(N{+}1) - \mathcal{L}_{\text{critic}}$ is therefore capped at $\log(N{+}1)$ nats, a known finite-sample ceiling of the InfoNCE bound~\citep{poole2019variational}. To evaluate the empowerment at an arbitrary state $s$ (rather than at a specific offline triple), $M$ joint samples $a \sim \pi_{\text{bc}}(\cdot\mid s)$, $s^+ \sim f_{\text{dyn}}(\cdot\mid s,a)$ are drawn and scored against fresh $\pi_{\text{bc}}$ negatives via the same bound and averaged. Because this substitutes $\pi_{\text{bc}}$ and $f_{\text{dyn}}$ for the true behavior policy and its discounted future-state distribution, the resulting estimate is the mutual information under the \emph{learned models} rather than the true environment, and errors in either model bias it.

\textbf{Offline adaptation.} $\pi_{\text{bc}}$ substitutes for the ability to roll out arbitrary actions from a live policy, and $f_{\text{dyn}}$ substitutes for rolling out $\pi_\beta$ over a discounted horizon; both are trained on the offline dataset with their own flow-matching losses, and no gradient from the critic flows into them.

\textbf{Hyperparameters.} Learning rate $3\times 10^{-4}$; batch size $256$; flow and critic hidden dims $(512,512,512)$ with layer normalization; $8$ Euler integration steps; $N=7$ negatives; $10^5$ steps of $\pi_{\text{bc}}$/$f_{\text{dyn}}$-only pretraining before the critic begins training.

\subsubsection{DV-Empowerment}
\label{sec:appendix-dv}

\textbf{Overview.} DV-Empowerment is architecturally identical to CF-FlowBC-Empowerment---the same $\pi_{\text{bc}}$, $f_{\text{dyn}}$, and scalar critic $T(s^+,a,s)$---but replaces the categorical InfoNCE bound with the Donsker--Varadhan (DV) lower bound~\citep{belghazi-mine}, and replaces the per-element action negatives with full $(\text{action}, \text{future-state})$ negative pairs drawn from an explicit product-of-marginals construction.

\textbf{Training objective.} Given a positive triple $(s,a,s^+)$, $N$ negative actions $a_1,\dots,a_N \overset{\text{iid}}{\sim} \pi_{\text{bc}}(\cdot\mid s)$ are sampled, each paired with its own simulated future state $s^+_i \sim f_{\text{dyn}}(\cdot\mid s,a_i)$. To obtain valid samples from the product of marginals $\pi_{\text{bc}}(a\mid s)\times f_{\text{dyn}}(s^+\mid s,a)$ rather than the joint, the future states are re-paired with a cyclic derangement $\sigma$ (e.g.\ $\sigma(i) = (i \bmod N) + 1$), so that $(a_i, s^+_{\sigma(i)})$ never uses the future state that $a_i$ itself generated. The DV objective is then
\begin{equation}
    J(s,a,s^+) = T(s^+,a,s) - \log\frac{1}{N}\sum_{i=1}^N e^{T(s^+_{\sigma(i)},\,a_i,\,s)}, \qquad \mathcal{L}_{\text{critic}} = -\E\big[J(s,a,s^+)\big],
\end{equation}
with reported estimate $\hat{I}(s^+;a\mid s) = \E[J(s,a,s^+)]$. As with CF-FlowBC-Empowerment, evaluating $\hat{\mathcal{E}}(s)$ at an arbitrary state draws $512$ fresh joint samples from $\pi_{\text{bc}}$ and $f_{\text{dyn}}$ and scores them against $512$ negatives built with the same derangement-based construction.

\textbf{Offline adaptation.} Identical to CF-FlowBC-Empowerment: $\pi_{\text{bc}}$ and $f_{\text{dyn}}$, both fit on the offline dataset, substitute for on-policy action and rollout sampling.

\textbf{Hyperparameters.} Identical to CF-FlowBC-Empowerment: learning rate $3\times10^{-4}$; batch size $256$; flow and critic hidden dims $(512,512,512)$ with layer normalization; $8$ Euler integration steps; $N=7$ negatives; $10^5$ steps of pretraining.

\subsubsection{DADS-Empowerment}
\label{sec:appendix-dads}

\textbf{Overview.} DADS-Empowerment is the only baseline that estimates the \emph{skill} empowerment $I(s^+;z\mid s)$ for a small, fixed-size discrete skill alphabet, matching the formalism used by our method. Rather than fixing a skill-labeling scheme first, as in DIAYN/DADS-style skill discovery~\citep{dads}, and only then measuring the resulting mutual information, DADS-Empowerment learns the skill assignment \emph{itself} so as to directly maximize the Barber--Agakov (BA) variational bound~\citep{barber-agakov} on offline $(s,s^+)$ pairs.

\textbf{Architecture.} Three networks operate on a fixed number $K$ of discrete, one-hot skills. An encoder $\rho(z\mid s,s^+)$ is an MLP that maps the concatenation of $s$ and the observed transition delta $s^+ - s$ to a softmax distribution over the $K$ skills; conditioning only on the endpoint pair (rather than any intermediate trajectory) ensures $z$ cannot encode information that never manifests in $s^+$ itself. A transition channel $q(s^+ \mid s, z)$ is a diagonal-Gaussian mixture with $8$ components conditioned on $s$ and a one-hot $z$, modeling the delta $s^+ - s$. A skill marginal $m(z\mid s)$ is an MLP over $s$ producing a softmax over the $K$ skills. There is no action-conditioned network of any kind, and no policy: the primitive action never appears in this baseline's training objective.

\textbf{Training objective.} For every offline pair $(s,s^+)$ and every skill $z \in \{1,\dots,K\}$ (the expectation over $z$ is computed exactly, not by Monte Carlo, since $K$ is small), define the BA bracket
\begin{equation}
    \text{bracket}(z; s,s^+) = \log q(s^+ \mid s,z) - \log\!\sum_{z'=1}^K m(z'\mid s)\, q(s^+\mid s,z'),
\end{equation}
where the log-sum-exp over $z'$ uses a skill marginal $m$. The encoder loss maximizes the bracket under its own soft skill assignment:
\begin{equation}
    \mathcal{L}_{\text{bracket}} = -\E_{(s,s^+)\sim\mathcal{D}}\Big[\textstyle\sum_{z=1}^K \rho(z\mid s,s^+)\,\text{bracket}(z;s,s^+)\Big],
\end{equation}
and the reported estimate is $\hat{I}(s^+;z\mid s) = -\mathcal{L}_{\text{bracket}}$. The skill marginal $m(z\mid s)$ is separately fit by cross-entropy toward the current (gradient-stopped) soft assignment,
\begin{equation}
    \mathcal{L}_{\text{marginal}} = -\E_{(s,s^+)\sim\mathcal{D}}\Big[\textstyle\sum_{z=1}^K \rho(z\mid s,s^+)\,\log m(z\mid s)\Big],
\end{equation}
so that $m$ tracks $\rho$'s marginal usage of each skill without corrupting $\rho$'s own gradient. A skill-usage regularizer, weighted by a coefficient $\lambda_{\text{kl}}$,
\begin{equation}
    \mathcal{L}_{\text{usage}} = -\log K - \frac{1}{K}\sum_{z=1}^{K} \log\bar\rho_z = D_{\mathrm{KL}}\big(\mathcal{U}_K \,\big\|\, \bar\rho\big), \qquad \bar\rho_z \triangleq \E_{(s,s^+)\sim\mathcal{D}}\big[\rho(z\mid s,s^+)\big],
\end{equation}
where $\mathcal{U}_K$ is the uniform distribution over the $K$ skills. This term penalizes deviation of the average skill usage from uniform, discouraging the assignment from collapsing onto a small subset of skills. The total loss is $\mathcal{L}_{\text{bracket}} + \mathcal{L}_{\text{marginal}} + \lambda_{\text{kl}} \mathcal{L}_{\text{usage}}$.

\textbf{Offline adaptation.} As with the other three baselines, $(s,s^+)$ pairs are drawn from the offline dataset; no skill-conditioned rollouts, dynamics-model rollouts, or EM alternation are used.

\textbf{Hyperparameters.} Learning rate $3\times10^{-4}$; batch size $1024$; mixture-density network with $8$ components; encoder, channel, and marginal networks all use hidden dims $(512,512,512)$ with layer normalization. We use $K = 15$ skills and $\lambda_{\text{kl}}=1$, including for the qualitative maps in \cref{sec:baseline-emp-maps}.

\end{document}